%% file: arxiv.tex
\documentclass[letterpaper, 10 pt, conference]{ieeeconf}
\usepackage{lipsum}
\usepackage{graphicx}
\usepackage{amsfonts}
\usepackage[T1]{fontenc}
\usepackage{amsmath}
\usepackage{amssymb}
\usepackage{multirow}
\usepackage[table]{xcolor}
\usepackage{cite}
\usepackage{booktabs}
\usepackage{bm}
\usepackage[font=small]{caption}
\usepackage{pifont}
\usepackage{makecell}
\usepackage{titlesec}
\usepackage{xspace}
\usepackage{hyperref}
\usepackage{cleveref}

\IEEEoverridecommandlockouts
\makeatletter
\newcommand{\teasercaption}[1]{%
  \refstepcounter{figure}%
  \@makecaption{\fnum@figure}{#1}%
}
\makeatother

\newcommand{\method}{SkillWeave}

\begin{document}    

\title{\LARGE \bf
SkillWeave: Weaving Heterogeneous Demonstrations into Long-Horizon Manipulation Skills
}

\author{
    Ryosei Tamura$^{1,2}$ \quad Xiaoxiang Dong$^{1}$ \quad Uksang Yoo$^{3}$ \quad Yuemin Mao$^{1}$ \\
    Romina Mir$^{1,4}$ \quad Jonathan Francis$^{1,4}$ \quad Jeffrey Ichnowski$^{1}$ \\
    \thanks{$^{1}$ Robotics Institute, Carnegie Mellon University, USA.}
    \thanks{$^{2}$ Department of Computer Science, Keio University, Japan.}
    \thanks{$^{3}$ College of Engineering, University of California, Berkeley, USA.}
    \thanks{$^{4}$ Bosch Center for Artificial Intelligence, Pittsburgh, PA, USA.}
}

\IEEEaftertitletext{
    \vspace{-4em}
    \begin{center}
    \includegraphics[width=\textwidth]{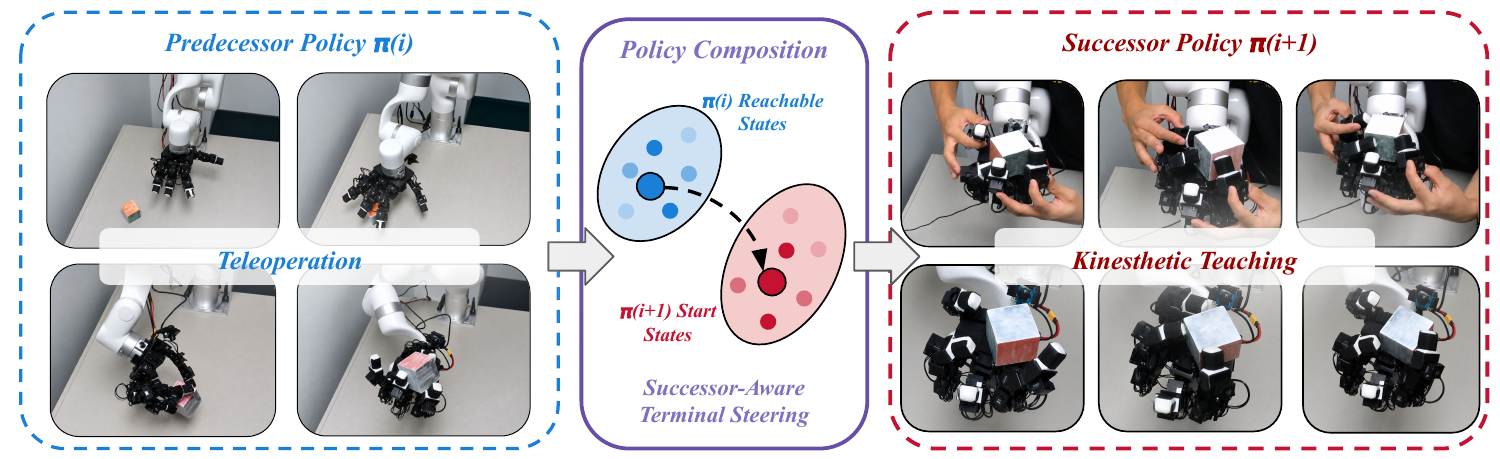}
    \vspace{-9pt}
    \teasercaption  {\footnotesize We propose \method{}, a framework that learns from multi-modal demonstration, where teleoperation is used for long-horizon coverage and free-space reaching (blue box), and kinesthetic teaching is used for precise contact-rich interaction (red box). Through policy composition with our proposed inference-time steering mechanism, \method{} is able to combine policies trained with individual sub-tasks and achieve long-horizon task success.}
    \label{fig:figure1}
    \vspace{-0.5em}
    \end{center}
}

\maketitle
    \begin{abstract} 
    Dexterous manipulation requires both large-scale task progression and precise contact-rich interaction, making it challenging to collect demonstrations that effectively support both regimes.
    We present \method{}, a heterogeneous demonstration framework for long-horizon dexterous manipulation that combines teleoperation for coarse reaching and transport with kinesthetic teaching for precise, contact-rich skills. 
    To address the visual mismatch introduced by the demonstrator's presence during kinesthetic data collection, we propose an object-mask-conditioned diffusion policy that uses offline object segmentation for training supervision and a lightweight learned mask predictor at deployment, avoiding online segmentation and image inpainting. 
    To mitigate distribution shift between independently trained sub-task policies, we introduce successor-aware terminal steering, which selects among actions sampled from the predecessor policy to guide the system toward states supported by the successor's demonstrated initial-state distribution.
    Across three real-world long-horizon tasks, \method{} achieves 27\% average end-to-end success. Mask-conditioned kinesthetic policies improve dexterous sub-task success to an average of 65\%, while successor-aware handoffs achieve an average composition efficiency of 87\%. 
    These results show that matching demonstration modality to interaction regime, explicitly addressing kinesthetic visual mismatch, and steering policy handoffs toward successor-supported states substantially improves long-horizon dexterous manipulation. Videos and code are available at \href{https://skillweave-authors.github.io/}{skillweave-authors.github.io}.

    \end{abstract}
    
\section{Introduction}
Dexterous manipulation combines high-dimensional control with repeated and often discontinuous changes in contact. Such behaviors have been studied through model-based control, reinforcement learning, and sim-to-real transfer \cite{kolbert2017contact,openai2020dexterous}. 
Demonstrations offer a complementary route by placing the learner directly on task-relevant state--action trajectories. They have been used both to accelerate reinforcement learning and to train expressive visuomotor policies through behavior cloning \cite{rajeswaran2018dexterous,chi2023diffusion}. 
Demonstration data can also be obtained or expanded through human videos and automated trajectory generation \cite{qin2022dexmv,chen2026videomanip,mandlekar2023mimicgen}. 
Regardless of its source, however, the resulting policy can learn only from the states and interactions represented in the data; demonstration collection therefore directly shapes the capabilities and failure modes of the learned behavior~\cite{lin2024datascaling}, and the specific \textit{type} of demonstration collection strategy used thus has a strong effect on the properties of the downstream policy \cite{yoo2025kinesoft}. Teleoperation captures task progression and spatial coverage, but retargeting and indirect contact feedback can limit precision during dexterous interaction \cite{naughton2024respilot,zhang2025kinedex}. 
On the other hand, kinesthetic teaching provides direct guidance through robot-feasible contact configurations, but introduces demonstrator occlusion into the recorded images \cite{chen2025dexforce,zhang2025kinedex}. These complementary limitations motivate our central hypothesis: under long-horizon dexterous manipulation, which alternates between free-space motion and contact-rich interaction, each phase should be demonstrated through the modality best suited to its interaction regime. We combine teleoperation demonstrations for task-level coverage with kinesthetic demonstrations at contact-rich phases.
Collecting each subtask separately also makes the data cheaper to gather, since a failed attempt costs one subtask rather than the whole episode.

However, data matched to the modulus of subtasks alone do not guaranty a successful composition. Each boundary introduced by the decomposition must be reached through two different processes: the demonstrator reaches it during data collection, whereas at deployment it must be reached by the preceding policy. Small prediction errors accumulate throughout that subtask, causing the predecessor's terminal state to drift away from the states represented in the demonstrations. The successor policy therefore begins from an out-of-distribution state and may fail despite succeeding in isolation; this mismatch can compound over a sequence of subtasks.
This terminal-to-initial distribution mismatch is a known limitation of naive skill composition ~\cite{lee2019composing,lee2022adversarial}.
Illustrated by \cref{fig:figure1}, we address this issue by inserting a short bridging motion between consecutive policies. The bridge guides the robot from the predecessor's realized terminal state toward the region of states represented by the successor's demonstrated initial state distribution.
  
Kinesthetic data also creates an observation mismatch: the demonstrator occludes the object during training but is absent during autonomous execution. Prior approaches remove visible embodiments through inpainting \cite{bahl2022human,chen2024mirage,zhang2025kinedex}, which requires reconstructing unobserved content near the interaction region. 
We instead use SAM~3 to generate object masks offline for the training demonstrations~\cite{carion2025sam3}. 
These masks supervise a learned mask predictor whose output conditions the diffusion policy during deployment.
This removes SAM~3 and image inpainting from the online control loop.
    
Our experiments test three claims: matching demonstration modality to interaction regime improves long-horizon, contact-rich manipulation; bridging into the successor's demonstrated initial states is necessary when independently learned behaviors are composed; and object-mask conditioning makes kinesthetic policy learning more robust to changes in background and scene.
    
In summary, this paper makes the following contributions:
\begin{itemize}
    \item We introduce a \textbf{multi-modal demonstration framework} that uses teleoperation for long-horizon coverage and free-space reaching, and kinesthetic
    teaching for precise contact-rich interaction.
    \item We develop \textbf{successor-aware terminal steering} to guide predecessor policies toward the successor's start-state distribution, reducing distribution shift at policy boundaries for reliable long-horizon composition.
    \item We develop an \textbf{object-mask-conditioned diffusion policy for kinesthetic demonstrations}. SAM~3 provides offline training masks, while a learned predictor supplies masks during deployment, improving robustness across backgrounds
    and scenes without online SAM~3 or image inpainting in the control loop.
\end{itemize}

 \section{Related Work}\label{sec:related_work}
    
\subsection{Demonstration Collection for Manipulation}
Robot learning from demonstration commonly relies on teleoperation, kinesthetic teaching, or observation of human behavior~\cite{argall2009survey}. These modalities expose the learner to different state--action distributions, making data coverage and diversity as important as demonstration count ~\cite{lin2024datascaling}.

Early dexterous teleoperation systems mapped observed human motion to robot actions through hand-pose estimation and kinematic retargeting~\cite{qin2022one}. 
More recent systems improve usability and data quality through immersive visual feedback, higher-rate tracking, and haptic interfaces \cite{ding2024bunny,iyer2025openteach,cheng2025television}. 
However, they retain an embodiment-matching problem: differences between human and robot kinematics can make reachable human poses inaccurate or infeasible for the robot.  
ResPilot reduces this mismatch through learned residual corrections, but remains limited by the underlying retargeting interface~\cite{naughton2024respilot}. 
These errors become particularly important during contact, where small configuration errors can change the resulting force or invalidate a grasp.

Kinesthetic teaching avoids retargeting by directly guiding the robot. DexForce extracts force-informed actions from kinesthetic demonstrations, but replays trajectories to obtain demonstrator-free images~\cite{chen2025dexforce}. Because replay is open loop, the object need not follow its original motion, so the recovered images may not correspond to the states in which the recorded actions were executed. KineDex instead preserves the original interaction and removes the demonstrator through segmentation and inpainting~\cite{zhang2025kinedex}, trading motion inconsistency for reconstruction error in occluded regions.

Existing systems therefore improve single modalities but fail to combine modalities tailored to distinct task phases. We address this by pairing teleoperation for long-horizon coverage with kinesthetic teaching for contact-rich corrections.
Finally, demonstrations are commonly converted into visuomotor policies via behavior cloning; for instance, Diffusion Policy models this as conditional denoising over action sequences, supporting multimodal prediction and receding-horizon execution~\cite{chi2023diffusion}.

\subsection{Skill Composition}
Long-horizon execution requires one skill's terminal states to fall within the next skill's initiation distribution. Prior work addresses this mismatch using learned transition policies \cite{lee2019composing}, terminal state regularization \cite{lee2022adversarial}, or feasibility-guided fine-tuning and switching \cite{chen2023sequential}. 
These methods require additional reinforcement-learning interaction and modify or augment the constituent policies.
Related methods instead keep a single policy near its training
distribution \cite{sun2025latentbarrier} or improve continuity between its action chunks \cite{he2026focalpolicy}, rather than connect independently trained skills. 
We use an explicit bridging motion to connect two policies while leaving their parameters unchanged.
    
\subsection{Visual Invariance in Demonstration Learning}
Early approaches improved visual robustness through augmentation or by removing the acting embodiment from images. GenAug uses generative image editing to vary objects and scenes, but does not address occlusion of task-relevant content ~\cite{chen2023genaug}. WHIRL inpaints humans and robots to construct an embodiment-invariant objective, while Mirage cross-paints a source robot over the target robot during deployment~\cite{bahl2022human,chen2024mirage}. These methods depend on reconstructing occluded pixels, which can introduce artifacts near contact regions.
    
Object-centric methods instead suppress irrelevant appearance through structured representations. VIOLA operates on object proposals, and GROOT uses object-centric 3D features~\cite{zhu2023viola,zhu2023groot}; however, both assume that task-relevant objects remain observable. More recently, SHADOW replaces robot pixels with composite masks, but requires calibrated robot geometry and test-time masking; its masks can also obscure robot--object contacts~\cite{lepert2025shadow}. KineDex applies inpainting to kinesthetic demonstrations, inheriting the difficulty of reconstructing objects hidden by the demonstrator~\cite{zhang2025kinedex}. Our method avoids image reconstruction and uses segmentation only as training-time supervision.

\section{Problem Statement}

We consider a long-horizon manipulation task $T$ composed of $K$ sequential sub-tasks,
\[
T = \left(T^{(1)}, T^{(2)}, \ldots, T^{(K)}\right).
\]
Each sub-task $T^{(i)}$ is demonstrated using a modality $m_i \in \{\textsc{tel}, \textsc{kin}\}$, where teleoperation is suited for coarse arm motions such as reaching and transport, while kinesthetic teaching enables precise, contact-rich manipulation. Demonstrations collected under $m_i$ form a dataset $\mathcal{D}^{(i)}$, from which we train a sub-task policy $\pi^{(i)}$.

Our goal is to compose the independently trained policies into a framework $\Pi$ that executes
\[
\pi^{(1)} \rightarrow \pi^{(2)} \rightarrow \cdots \rightarrow \pi^{(K)}
\]
to complete $T$. This introduces two challenges. First, kinesthetic demonstrations contain human-hand occlusions that are absent during autonomous execution, creating a demonstration-to-deployment visual mismatch. Second, independent policy training does not guarantee that the terminal state of $\pi^{(i)}$ is suitable for initializing $\pi^{(i+1)}$, making policy handoff a potential failure mode.

We therefore study how to learn from heterogeneous demonstration modalities while addressing modality-specific visual mismatch and reliably composing the resulting sub-task policies into long-horizon behaviors.

\begin{figure}[t]
    \centering
    \includegraphics[width=\columnwidth]{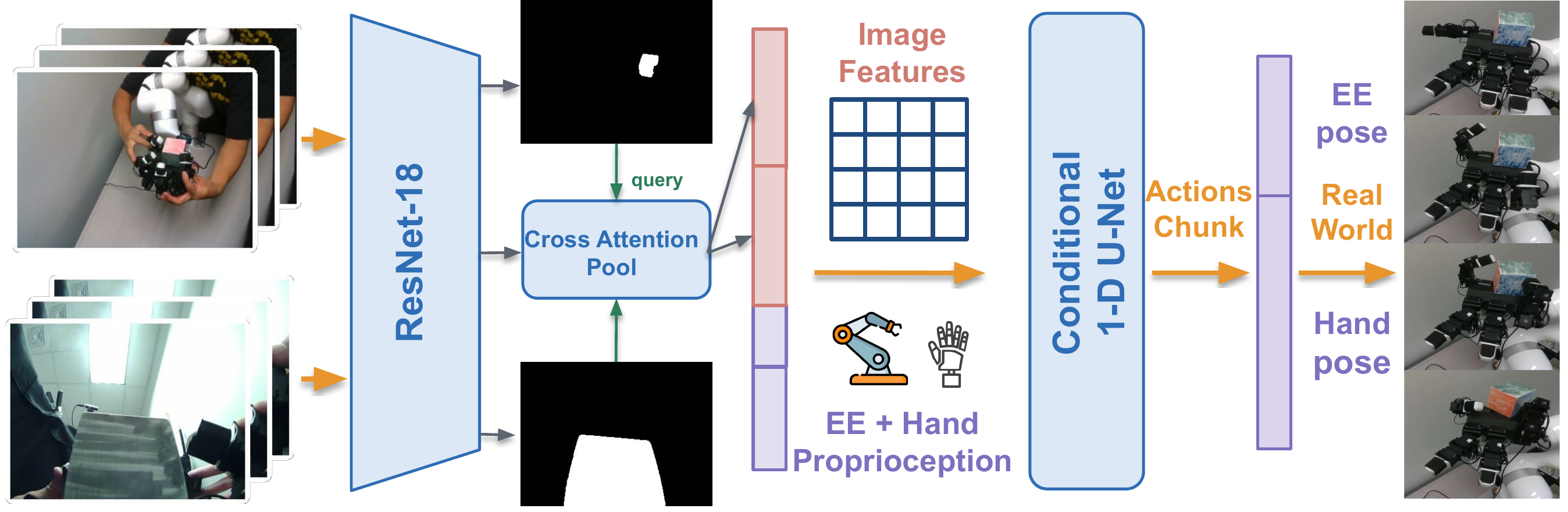}
    \caption[]
    {\footnotesize \textbf{ Mask-conditioned diffusion policy architecture.} External- and wrist-camera RGB observations are encoded by ResNet-18, and object masks guide cross-attention pooling over the resulting spatial feature maps. The pooled visual features are concatenated with end-effector and hand proprioception to condition a 1-D U-Net diffusion model that predicts robot action chunks. Training uses offline SAM~3 masks, while deployment uses learned mask predictors.}
    \vspace{-1.5em}
    \label{fig:masked-dp-arch}
\end{figure}

\begin{figure*}[t]
    \centering
    \includegraphics[width=\textwidth]{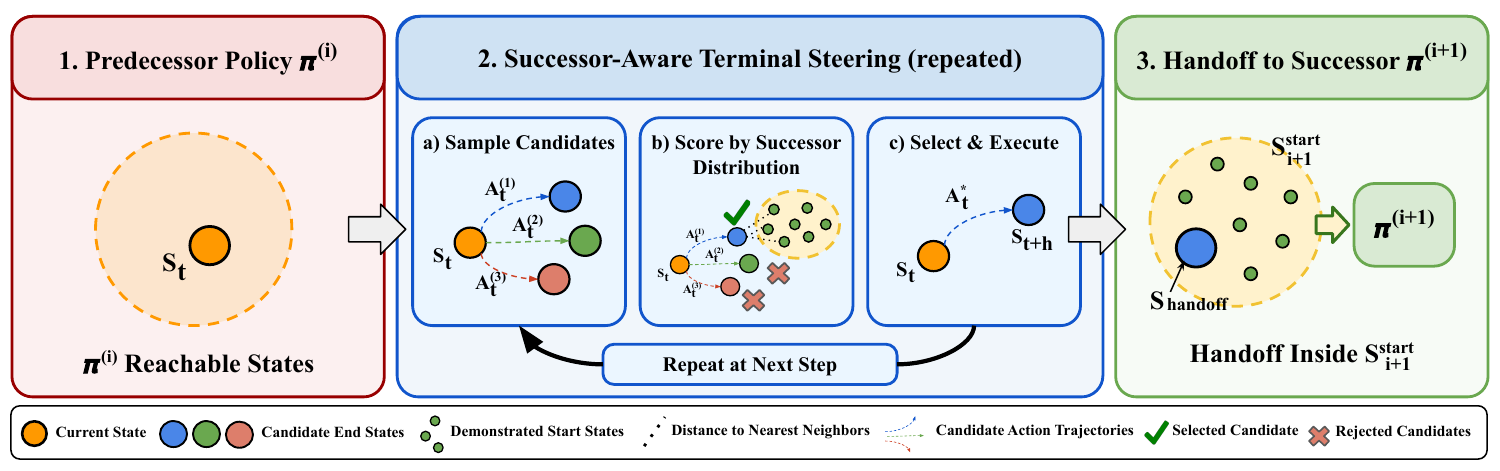}
    \caption[]
    {\footnotesize \textbf{Overview of the policy composition and handoff framework. }A predecessor policy $\pi^{(i)}$ executes its sub-task until entering the terminal phase. To facilitate a reliable transition to the independently trained successor policy $\pi^{(i+1)}$, we repeatedly sample candidate action chunks from $\pi^{(i)}$, evaluate their predicted terminal states according to proximity to the successor’s demonstrated start-state distribution $\mathcal{S}_{i+1}^{\mathrm{start}}$, and execute the best candidate. This closed-loop steering procedure moves the system toward a state supported by the successor’s training distribution, after which control is handed off to $\pi^{(i+1)}$. The approach reduces distribution shift at policy boundaries without requiring additional transition demonstrations or a separately trained policy.}
    \vspace{-1.5em}
    \label{fig:policy-composition-arch}
\end{figure*}

\section{Method}

\subsection{Sub-Task Policies}
\label{sec:policies}

For each sub-task $T^{(i)}$, we train a diffusion policy~\cite{chi2023diffusion} $\pi^{(i)}$ from its corresponding demonstration dataset $\mathcal{D}^{(i)}$. The observation space consists of
\[
\mathbf{o}_t =
\left(\mathbf{I}^{\mathrm{ext}}_t,\,
      \mathbf{I}^{\mathrm{wrist}}_t,\,
      \mathbf{q}_t\right),
\]
where $\mathbf{I}^{\mathrm{ext}}_t$ and $\mathbf{I}^{\mathrm{wrist}}_t$ are RGB observations from a third-person workspace camera and a wrist-mounted camera facing the robot palm, respectively. The proprioceptive state $\mathbf{q}_t$ contains the robot's end-effector Cartesian pose, arm joint angles, and hand joint angles.

Each camera observation is encoded by a ResNet-18 visual encoder $\phi_v$. The resulting visual features are combined with $\mathbf{q}_t$ to condition a 1D U-Net $\epsilon_\theta$ that denoises an action chunk. The policy is trained using the diffusion denoising objective:
\begin{align}
\mathcal{L}_{\mathrm{DP}}
=
\mathbb{E}_{\mathbf{a}_0,\epsilon,t}
\left[ \left|
    \epsilon - \epsilon_\theta \left(
        \mathbf{a}_t,t \mid \phi_1(\mathbf{I}^1), \phi_2(\mathbf{I}^2), \mathbf{q}
    \right) \right|_2^2 \right].\label{eq:dp}
\end{align}

\textbf{Teleoperation Policies.}
Teleoperated demonstrations contain only the robot and objects in the workspace and therefore exhibit little visual mismatch between demonstration and deployment. Following standard image-based diffusion policies, we aggregate each encoder feature map using spatial softmax~\cite{levine2016visuomotor} and concatenate visual representations with robot proprioception to condition the diffusion policy. 

\textbf{Mask-Conditioned Kinesthetic Policies.}
Kinesthetic teaching introduces a different observation distribution: the demonstrator's hand and arm are visible during data collection but absent during autonomous execution. This occlusion can cause the visual encoder to attend to the demonstrator rather than the manipulated object, resulting in a substantial demonstration-to-deployment visual mismatch.

Rather than removing or inpainting the demonstrator, we retain the original RGB observations and condition visual feature aggregation on a mask of the manipulated object. The architecture of mask-conditioned diffusion policy is shown in~\cref{fig:masked-dp-arch}. During training, object masks are generated offline using SAM~3~\cite{carion2025sam3} for each camera view.

Specifically, let $F_v \in \mathbb{R}^{C \times h \times w}$ denote the visual feature map for camera view $v$, and let $m_v \in [0,1]^{h \times w}$ denote the corresponding object mask downsampled to the feature resolution. We construct an object-conditioned query by pooling features within the masked region:
\begin{align}
    q_v = \frac{\sum_{k} m_{v,k} F_{v,k}}{\sum_{k} m_{v,k}},
    \qquad
    z_v = \mathrm{AttnPool}(F_v,\ q_v),
    \label{eq:pool}
\end{align}
where $z_v$ is the resulting visual representation. When the object is not visible in a camera view, we fall back to the mean feature as the attention query.

To further encourage the visual representation to focus on the manipulated object, we supervise the spatial attention distribution using the object mask. Let $a_v$ denote the attention weights over the feature grid and $\bar m_v = m_v / \lVert m_v \rVert_1$ the normalized object mask. We define
\begin{align}
    \mathcal{L}_{\mathrm{attn}} &= \sum_v \mathbb{E} \left[
        -\sum_k \bar m_{v,k}\log a_{v,k} \right], \\
    \mathcal{L} &= \mathcal{L}_{\mathrm{DP}} + \lambda \mathcal{L}_{\mathrm{attn}},
    \label{eq:attn}
\end{align}
where $\lambda$ controls the strength of the attention supervision. Views in which the manipulated object is not visible do not contribute to $\mathcal{L}_{\mathrm{attn}}$. 

At deployment, SAM~3 is not required online. Instead, we train a lightweight U-Net for each camera view to predict the manipulated-object mask directly from the RGB observation. The predicted masks replace the offline SAM~3 masks and condition the policy in the same manner as during training. This design preserves the original RGB observations while providing an object-centric visual representation that is robust to the disappearance of the human demonstrator.

\subsection{Policy Composition and Successor-Aware Handoff}
\label{sec:policy-composition}

\method{} decomposes long-horizon tasks into sub-tasks with independently trained policies using optimal modalities. However, independent training does not guarantee compatibility between consecutive policies: terminal states of policy $\pi^{(i)}$ may deviate from demonstrated initial states of successor $\pi^{(i+1)}$, creating out-of-distribution initialization failures during policy switching.

We address this using \textit{successor-aware terminal steering}, modifying predecessor termination instead of adding connecting policies. Near sub-task completion, $\pi^{(i)}$ enters a steering phase that guides it toward high-support start states of $\pi^{(i+1)}$. 
By exploiting diffusion policy stochasticity, we sample multiple action chunks per step and select the candidate closest to the successor's demonstrated start states, steering the system into a successful handoff region.

\textbf{Successor start-state distribution.}
For each successor policy $\pi^{(i+1)}$, we extract the first $H_s$ states from its $N$ training demonstrations and construct
\[
\mathcal{S}_{i+1}^{\mathrm{start}}
=
\left\{
s_{i+1,t}^{(n)}
\mid
n=1,\ldots,N,\;
t=1,\ldots,H_s
\right\}.
\]

These states provide an empirical representation of the region from which the successor policy was trained to begin its behavior. We represent each handoff state using the robot configuration relevant to policy execution, including the wrist pose, arm joint configuration, and hand joint configuration. Each component is normalized before computing distances to account for differences in scale.

Given a state $s$, we estimate its compatibility with the successor using its $K$ nearest neighbors in $\mathcal{S}_{i+1}^{\mathrm{start}}$:
\[
d_{i+1}(s)
=
\frac{1}{K}
\sum_{s' \in
\mathrm{kNN}_K(s,\mathcal{S}_{i+1}^{\mathrm{start}})}
d(s,s'),
\]

where $d(\cdot,\cdot)$ denotes the normalized state-space distance. A smaller $d_{i+1}(s)$ indicates greater compatibility with the successor's demonstrated initial-state distribution.

\input{floats/table_task_summary}

\textbf{Iterative terminal steering.}
During normal execution, $\pi^{(i)}$ operates without modification. Once the sub-task enters its terminal phase, we perform successor-aware steering over multiple replanning steps. At steering step $t$, we sample $L$ candidate action chunks
\[
A_t^{(1)},\ldots,A_t^{(L)}
\sim
\pi^{(i)}(\cdot\mid o_t).
\]

Each candidate produces a prospective endpoint $\hat{s}_{t+H}^{(l)}$. We evaluate these endpoints against the successor start-state distribution and select
\[
A_t^*
=
\arg\min_{A_t^{(l)}}
d_{i+1}\!\left(\hat{s}_{t+H}^{(l)}\right).
\]

After executing the selected action chunk, the robot receives a new observation and repeats the candidate-sampling and selection procedure. Thus, terminal steering forms a closed-loop process,
\[
s_t
\xrightarrow{A_t^*}
s_{t+H}
\xrightarrow{A_{t+H}^*}
s_{t+2H}
\xrightarrow{\;\cdots\;}
s_{\mathrm{handoff}},
\]
where each selected action moves the system toward greater support under the successor's start-state distribution. The steering phase terminates when $d_{i+1}(s_{\mathrm{handoff}}) < \epsilon$, indicating that the current state is sufficiently close to the demonstrated start states of $\pi^{(i+1)}$, after which control is transferred to the successor policy. A maximum steering horizon is additionally imposed to prevent indefinite steering when the successor distribution cannot be reached.

Importantly, successor-aware steering does not construct an arbitrary trajectory between two policies. Every candidate action is sampled directly from $\pi^{(i)}$, so the system remains within behaviors supported by the predecessor while progressively moving toward states familiar to the successor. This distinguishes our approach from explicit interpolation or a separately learned transition controller: the transition is realized among the predecessor's own action proposals.

Conceptually, the procedure steers the terminal-state distribution of the predecessor toward the demonstrated start-state distribution of the successor,
\[
p_i^{\mathrm{term}}(s)
\;\longrightarrow\;
p_{i+1}^{\mathrm{start}}(s).
\]

It requires neither additional transition demonstrations nor a separately trained transition policy, while leaving $\pi^{(i+1)}$ unchanged. This enables independently trained heterogeneous sub-task policies to be composed while directly reducing the distribution shift encountered at policy boundaries.

\section{Experiments}\label{sec:results}

We evaluate \method{} on three real-world, long-horizon manipulation tasks. Our experiments are designed to answer the following questions:
\begin{enumerate}
    \item \textbf{Skill learning from kinesthetic demonstrations:} Can the proposed mask-conditioned diffusion policy learn skills from human kinesthetic demonstrations and autonomously execute them at test time, without the demonstrator's hand present in the observation?
    \item \textbf{Long-horizon skill composition:} Given trained policies for individual sub-tasks, can the proposed framework effectively compose these sub-policies to achieve robust long-horizon task execution?
\end{enumerate}

\subsection{Demonstration Collection}
We collect demonstrations using a 7-DoF xArm7 robot arm equipped with a 16-DoF LEAP Hand. Visual observations are captured by a third-person RGB camera overlooking the workspace and a wrist-mounted RGB camera directed toward palm. 

\begin{figure}[t]
    \centering
    \includegraphics[width=\columnwidth]{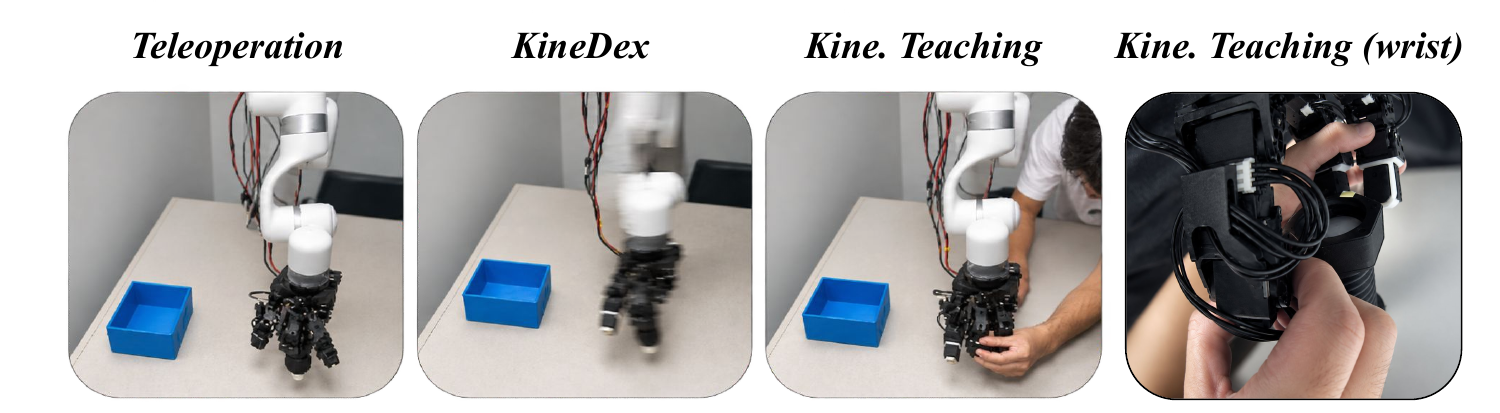}
     \caption[]
    {\footnotesize \textbf{Example RGB observations for policies.} Raw observations during teleoperation and kinesthetic teaching are directly fed to policies; KineDex inpaints the image to align with inference-time observation, but often blurs out robot hand and generates hallucinated parts.}
    \vspace{-1.5em}
    \label{fig:rw-kine-demo}
\end{figure}

For each sub-task, we collect 100 demonstrations independently using either teleoperation or kinesthetic teaching, depending on the required motion modality. During teleoperation, the operator controls the arm using a 6-DoF SpaceMouse. The operator's finger movements are captured by a camera, tracked using MediaPipe, and retargeted to the robot hand through joint-angle mapping. During kinesthetic teaching, the arm operates in drag/teach mode, while the operator manually positions the robot's fingers under a compliant controller that yields to manual adjustments and holds the resulting poses. Example observations during kinesthetic teaching are shown in~\cref{fig:rw-kine-demo}. 

Object positions are randomized across episodes to encourage spatial generalization. 
To train separate sub-task policies, we record synchronized RGB observations from both cameras and robot proprioceptive states, including the robot's 6D end-effector Cartesian pose, 7D arm joint angles, and 16D hand joint angles, at 30~Hz.

\subsection{Task Descriptions}

We evaluate our framework on three long-horizon manipulation tasks that require composing skills demonstrated through teleoperation and kinesthetic teaching.

\textbf{Cube Reorientation and In-Hand Rotation.}\label{task:cube}
Reaching, grasping, and wrist-reorientation to place the cube on the hand are demonstrated via teleoperation, whereas dexterous in-hand rotation is demonstrated via kinesthetic teaching.

\textbf{Nut Removal and Storage.}\label{task:nut}
Reaching for a threaded nut and placing it into a container are demonstrated via teleoperation, whereas dexterous unscrewing from the bolt is demonstrated via kinesthetic teaching.

\textbf{Kettle Preparation and Placement.}\label{task:kettle}
Reaching for the kettle and placing it underneath a faucet are demonstrated via teleoperation, whereas handle grasping and pressing the lid release button are demonstrated via kinesthetic teaching.

The decomposition of each long-horizon task into sub-tasks is summarized in~\cref{tab:task-summary}, and visualized in~\cref{fig:\method{}-results}.

\subsection{Evaluation Metrics}

We report individual sub-task success $S_i^{(j)}$ and end-to-end task success $S_i$. To quantify losses introduced by policy composition, we compute the expected success from isolated sub-task performance as $S_i^{*}=\prod_{j=1}^{M_i}S_i^{(j)}$, and define the composition efficiency as $C_i=S_i/S_i^{*}$. Values near $1$ indicate that policy handoffs introduce little additional degradation beyond the isolated sub-task failure rates.

\subsection{Baselines} 

We compare \method{} against three baselines that use a single demonstration modality or an alternative approach for handling the visual mismatch introduced by kinesthetic teaching. All baselines do not have explicit steering composition during long-horizon executions. 

\textbf{Teleoperation Only.}
All sub-task policies are trained using demonstrations collected through teleoperation. Note that we collect demonstrations using teleoperation regardless of success or failure, which we expect most demos to fail for complex dexterous tasks. This baseline evaluates whether teleoperation alone is sufficient for both coarse arm motions and dexterous hand manipulation, and quantifies the benefit of introducing kinesthetic demonstrations for skills that are difficult to teleoperate. 

\textbf{Kinesthetic Teaching Only.}
All sub-task policies are trained using kinesthetic demonstrations. This baseline evaluates whether a single kinesthetic teaching modality can support the complete task. In particular, it exposes the visual distribution shift caused by the demonstrator's hand and arm being present during demonstration collection but absent during autonomous execution.

\textbf{KineDex~\cite{zhang2025kinedex}.}
We additionally compare against KineDex, a state-of-the-art framework for learning dexterous manipulation from kinesthetic demonstrations. KineDex addresses the visual mismatch between demonstration and deployment by segmenting the human hand from kinesthetic demonstration frames and inpainting the corresponding image regions. This baseline represents an alternative strategy for resolving the demonstration--deployment visual gap: modifying the demonstration observations to resemble inference-time observations, rather than explicitly conditioning the policy on object masks as in \method{}.

\section{Results}

\input{floats/table_main_results}

\input{floats/table_long_horizon_results}

\begin{figure*}[t]
    \centering
    \includegraphics[width=\textwidth]{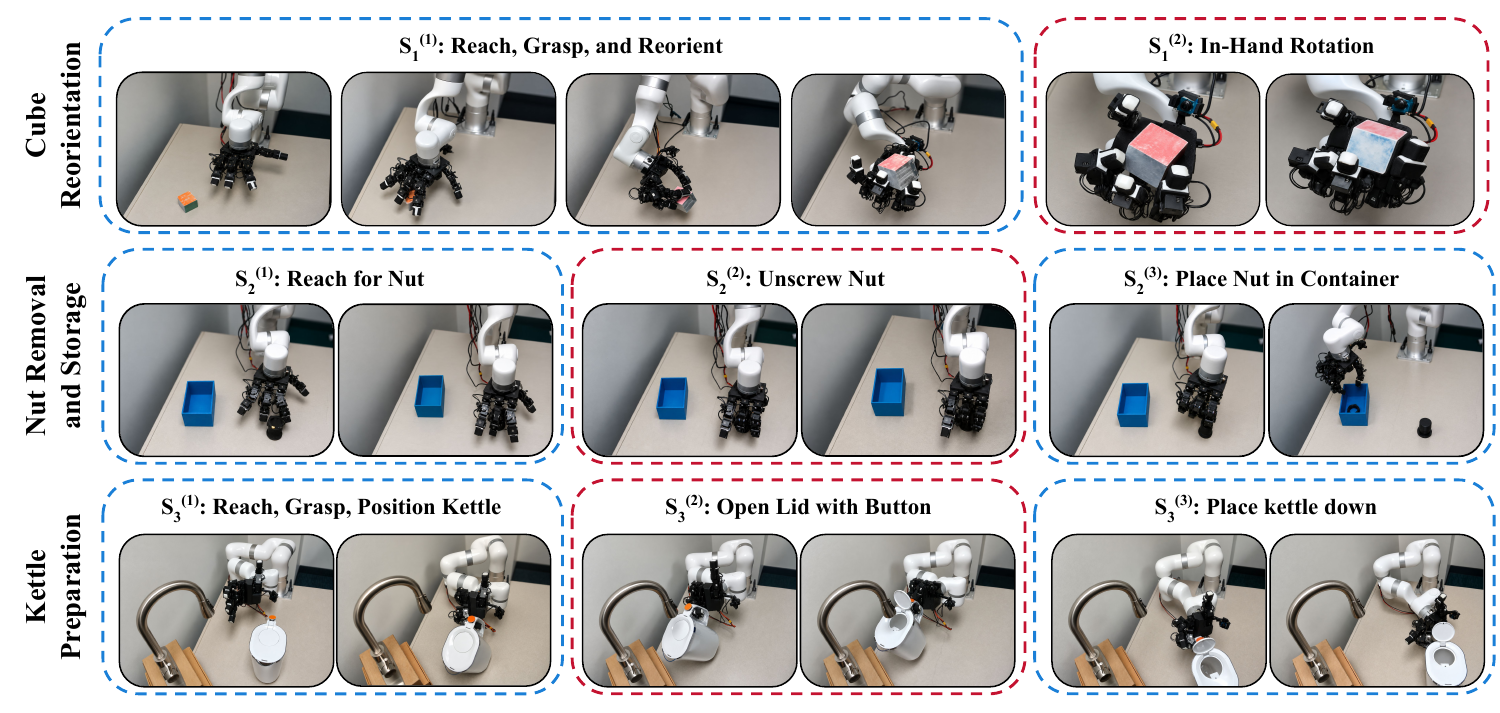}
     \caption[]
    {\footnotesize  \textbf{\method{} long-horizon visualizations.} From top to bottom: Cube Reorientation, Nut Removal and Storage, and Kettle Preparation. Blue shows frames for policies trained with teleoperation, while red shows frames for mask-conditioned kinesthetic teaching policies. }
    \label{fig:\method{}-results}
    \vspace{-1.45em}
\end{figure*}

\begin{figure}[t]
    \centering
    \includegraphics[width=\columnwidth]{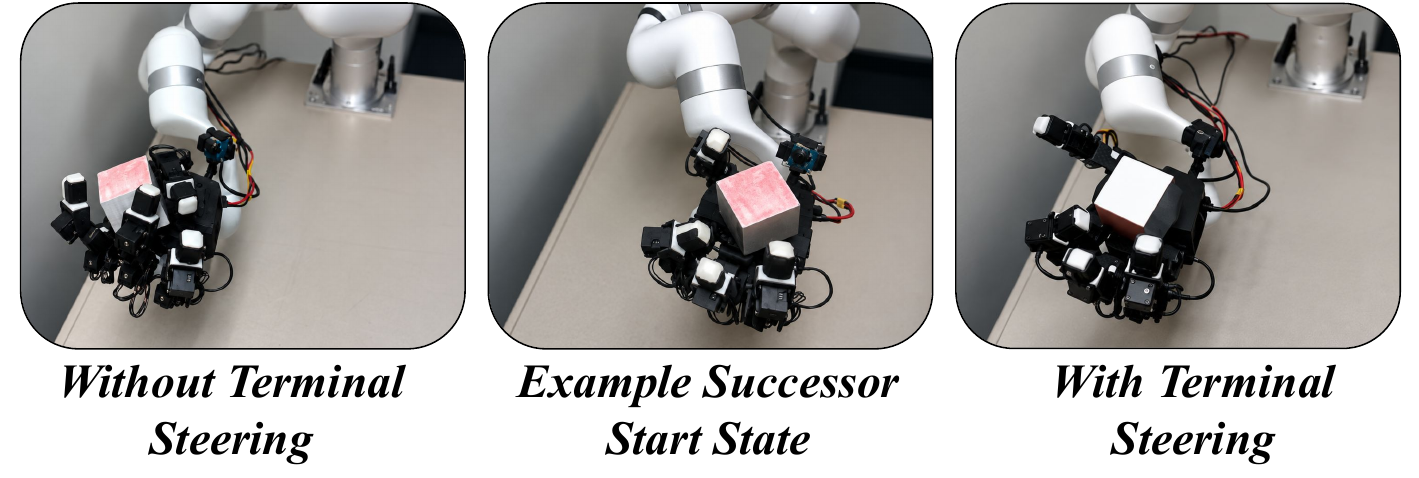}
     \caption[]
    {\footnotesize \textbf{Comparison between using successor-aware terminal steering and not using terminal steering.} Without terminal steering (left), predecessor policy ended at finger joint states significantly different from successor policy start joint states (middle); In comparison, with terminal steering (right), predecessor policy ended at finger joint states closer to successor policy start joint states.}
    \vspace{-1.6em}
    \label{fig:steering-rw}
\end{figure}

\subsubsection*{Sub-Task Policy Evaluation:}
We first evaluate each learned sub-task policy independently. At the sub-task level, we compare teleoperation, kinesthetic teaching without masking, KineDex~\cite{zhang2025kinedex}, and our mask-conditioned kinesthetic policy. This evaluates both the suitability of different demonstration modalities for different manipulation skills and the effectiveness of mask conditioning for learning from kinesthetic demonstrations.

Each policy is evaluated over $N=20$ rollouts with randomized object configurations, and the resulting success rates are reported in~\cref{tab:main-results}.

For sub-tasks dominated by arm-level reaching, reorientation, and placement, teleoperation achieves an average success rate of $64\%$. In comparison, raw kinesthetic teaching and KineDex achieve averaged only $4.5\%$ on the same set of sub-tasks. These results characterize the relative suitability of teleoperation for coarse, workspace-level manipulation. 

A different trend emerges for sub-tasks requiring precise dexterous hand control. Across in-hand cube rotation, nut unscrewing, kettle-handle grasping, and button pressing, teleoperation achieves an average success rate of $5\%$, while mask-conditioned kinesthetic teaching achieves $65\%$. These results demonstrate the advantage of directly guiding the robot hand through the desired configurations for skills that require coordinated finger-level control.

We next isolate the effect of visual processing for kinesthetic demonstrations. When trained directly on raw kinesthetic observations without masking, policies achieve an average success rate of $10\%$ across the dexterous sub-tasks. KineDex, which removes the demonstrator's hand through segmentation and image inpainting, achieves $20\%$, while our mask-conditioned policy achieves $65\%$. All policies are evaluated autonomously without the demonstrator's hand present. The improvement from raw kinesthetic observations to mask conditioning indicates that explicitly accounting for the demonstration-time occlusion is important for bridging the visual distribution shift between kinesthetic demonstration and autonomous execution. Furthermore, the comparison with KineDex evaluates whether providing explicit object-centric mask information is more effective than reconstructing the occluded image regions through inpainting.

Taken together, the sub-task results support the use of heterogeneous demonstration modalities: teleoperation is well suited for coarse arm-level motions, whereas kinesthetic teaching provides an advantage for skills requiring precise dexterous control. In addition, mask conditioning enables policies trained from kinesthetic demonstrations to execute autonomously despite the absence of the demonstrator.

\subsubsection*{Long-Horizon Task Evaluation:}
We next evaluate whether the learned sub-task policies can be composed to solve complete long-horizon tasks. For each task, we perform $N=20$ end-to-end rollouts with randomized initial object configurations. We compare \method{} against three single-strategy systems: Teleoperation Only, Mask-Conditioned Kinesthetic Only, and KineDex Only, as well as an ablation on multi-modal demonstrations without any composition steering (SkillWeave without steering). Results are reported in~\cref{tab:main-results}, and representative \method{} rollouts are visualized in~\cref{fig:\method{}-results}.

\method{} achieves end-to-end success rates of $45\%$, $15\%$, and $20\%$ on Cube Reorientation, Nut Removal and Storage, and Kettle Preparation, respectively, corresponding to an average success rate of $27\%$. In comparison, Teleoperation Only, Mask-Conditioned Kinesthetic Only, and KineDex alone, cannot get a single success. The composition-efficiency results in~\cref{tab:long-horizon-results} further demonstrate the importance of terminal steering. Naively composing the same sub-task policies without steering results in a substantial degradation in end-to-end success, indicating that strong isolated sub-task performance alone does not guarantee reliable long-horizon execution. Terminal steering mitigates this degradation by guiding policy handoffs toward states compatible with the successor policy. 
A qualitative comparison of real-world rollouts with and without terminal steering is shown in~\cref{fig:steering-rw}.

These results highlight both the limitations of applying a single demonstration strategy uniformly across heterogeneous manipulation skills and the importance of reliable policy handoffs. Teleoperation Only is limited by precise, dexterous manipulation stages, while approaches relying exclusively on kinesthetic demonstrations exhibit lower reliability for large-workspace reaching and placement motions. \method{} instead exploits the complementary strengths of the two modalities: teleoperation is used for coarse arm-level manipulation, while mask-conditioned kinesthetic teaching is used for precise hand-level manipulation. Successor-aware terminal steering then composes these independently learned policies by reducing distribution shift at policy boundaries. Together, these components allow each sub-task to be learned using a demonstration modality suited to its control requirements and reliably composed into long-horizon behaviors.

\section{Conclusions and Future Works}  
    
We present \method{}, a heterogeneous demonstration framework for long-horizon dexterous manipulation that combines teleoperation for coarse arm motions with kinesthetic teaching for precise, contact-rich interactions. To enable robust autonomous execution, we introduced object-mask-conditioned diffusion policies to mitigate the visual mismatch in kinesthetic demonstrations, as well as bridging motions to address policy handoff failures during skill composition. Experiments on three real-world long-horizon tasks demonstrate the complementary strengths of the two demonstration modalities and show that the resulting heterogeneous policies can be effectively composed for end-to-end execution. Future work could automate task decomposition and demonstration modalities decisions by jointly learning these from data, as well as to reduce the reliance on explicit object masks and extend the framework to larger skill libraries, novel objects, and more diverse environments.
    
\vspace{-0.8em}

\bibliographystyle{IEEEtran}
\bibliography{references}

\end{document}

%% file: floats/table_task_summary.tex
\begin{table*}[t]
    \centering
    \caption{
        Long-horizon task decomposition and sub-task success criteria.
        $S_i^{(j)}$ is determined by the corresponding criterion.
    }
    \label{tab:task-summary}

    \small
    \setlength{\tabcolsep}{6pt}
    \renewcommand{\arraystretch}{1.05}

    \begin{tabular}{@{}lllp{0.43\textwidth}@{}}
        \toprule
        \textbf{Task ($i$)}
        & \textbf{$j$}
        & \textbf{Sub-task}
        & \textbf{Success Criterion} \\
        \midrule

        \multirow{2}{*}{Cube Reorientation}
        & 1 & Reach, grasp, and reorient cube
        & Cube securely grasped without dropping \\

        & 2 & Rotate cube in-hand
        & Cube rotated by at least $90^\circ$ \\

        \midrule

        \multirow{3}{*}{Nut Removal \& Storage}
        & 1 & Reach for nut
        & Hand positioned over the nut \\

        & 2 & Unscrew nut
        & Nut fully detached from screw \\

        & 3 & Place nut in container
        & Nut released inside the container \\

        \midrule

        \multirow{3}{*}{Kettle Preparation}
        & 1 & Reach, grasp, and position kettle
        & Kettle grasped by handle and positioned under faucet \\

        & 2 & Open lid using release button
        & Release button pressed and lid opened \\

        & 3 & Place kettle down
        & Kettle released and stable on table \\

        \bottomrule
    \end{tabular}
    \vspace{-1.5em}
\end{table*}

%% file: floats/table_main_results.tex
\begin{table}[t]
    \centering
    \caption{
        \textbf{Sub-task and long-horizon success rates.}
        (a) Success rates of individual sub-task policies across demonstration
        and visual-processing strategies.
        (b) End-to-end success rates of long-horizon systems.
        All results are reported over 20 trials.
    }
    \label{tab:main-results}

    \small
    \setlength{\tabcolsep}{3pt}
    \renewcommand{\arraystretch}{1.15}

    \textbf{(a) Sub-task Success} \\[3pt]

    \resizebox{\columnwidth}{!}{
    \begin{tabular}{lcccc}
        \toprule
        \textbf{Sub-task}
        & \textbf{Teleop.}
        & \textbf{Kin.}
        & \textbf{KineDex}
        & \textbf{Mask-Kin.} \\
        \midrule

        Reach, grasp, \& reorient
        & 70\% & 0\% & 0\% & --- \\

        In-hand rotation
        & 10\% & 30\% & 0\% & \textbf{70\%} \\

        \midrule

        Reach for nut
        & 85\% & 0\% & 40\% & --- \\

        Unscrew nut
        & 5\% & 0\% & 60\% & \textbf{75\%} \\

        Place nut in container
        & 30\% & 0\% & 5\% & --- \\

        \midrule

        Reach, grasp, \& position kettle
        & 80\% & 0\% & 0\% & --- \\

        Open lid with button
        & 0\% & 0\% & 0\% & \textbf{50\%} \\

        Place kettle down
        & 55\% & 0\% & 0\% & --- \\

        \bottomrule
    \end{tabular}}

    \vspace{8pt}

    \textbf{(b) Long-Horizon Success} \\[3pt]

    \resizebox{\columnwidth}{!}{
    \begin{tabular}{lccccc}
        \toprule
        \textbf{Task}
        & \textbf{Teleop.}
        & \textbf{Kin.}
        & \textbf{KineDex}
        & \textbf{\method{}}
        & \textbf{\method{}} \\
        & & & &
        \textbf{w/o Steering}
        & \\
        \midrule
    
        Cube Reorientation
        & 0\% & 0\% & 0\% & 10\% & \textbf{45\%} \\
    
        Nut Removal \& Storage
        & 0\% & 0\% & 0\% & 0\% & \textbf{15\%} \\
    
        Kettle Preparation
        & 0\% & 0\% & 0\% & 5\% & \textbf{20\%} \\

        \bottomrule
    \end{tabular}}
\end{table}

%% file: floats/table_long_horizon_results.tex
\begin{table}[t]
    \centering
    \caption{
        Long-horizon task performance and composition efficiency. Note Kinesthetic-only results are omitted because $S^*=0$ for all tasks, making $C$ undefined.
    }
    \label{tab:long-horizon-results}

    \footnotesize
    \setlength{\tabcolsep}{2.5pt}
    \renewcommand{\arraystretch}{1.08}

    \begin{tabular}{@{}l*{4}{ccc}@{}}
        \toprule
        & \multicolumn{3}{c}{\textbf{Teleop}}
        & \multicolumn{3}{c}{\textbf{KineDex}}
        & \multicolumn{3}{c}{
            \makecell{\textbf{\method{}} \\ \textbf{w/o Steering}}}
        & \multicolumn{3}{c}{\textbf{\method{}}} \\
        \cmidrule(lr){2-4}
        \cmidrule(lr){5-7}
        \cmidrule(lr){8-10}
        \cmidrule(lr){11-13}

        \textbf{Task}
        & $S$ & $S^*$ & $C$
        & $S$ & $S^*$ & $C$
        & $S$ & $S^*$ & $C$
        & $S$ & $S^*$ & $C$ \\
        \midrule

        Task 1
        & 0\% & 7\% & 0\%
        & 0\% & 0\% & --
        & 10\% & 49\% & 20\%
        & 45\% & 49\% & $\mathbf{92\%}$ \\

        Task 2
        & 0\% & 1\% & 0\%
        & 0\% & 1\% & 0\%
        & 0\% & 19\% & 0\%
        & 15\% & 19\% & $\mathbf{79\%}$ \\

        Task 3
        & 0\% & 0\% & --
        & 0\% & 0\% & --
        &  5\% & 22\% & 23\%
        & 20\% & 22\% & $\mathbf{91\%}$ \\

        \midrule
        \textbf{Avg.}
        & 0\% & 3\% & 0\%
        & 0\% & 1\% & 0\%
        & 5\% & 30\% & 14\%
        & 27\% & 30\% & $\mathbf{87\%}$ \\

        \bottomrule
    \end{tabular}
    \vspace{-1.5em}
\end{table}